# FRAC-Q-Learning: A Reinforcement Learning with Boredom Avoidance Processes for Social Robots

Akinari Onishi[1*]

1* Department of Electronic Systems Engineering, National Institute of Technology, Kagawa College, 551, Kohda, Takuma-cho, Mitoyo-City, 769-1192, Kagawa, Japan

*Corresponding author. E-mail: onishi-a@es.kagawa-nct.ac.jp

## Abstract

The reinforcement learning algorithms have often been applied to social robots. However, most reinforcement learning algorithms were not optimized for the use of social robots, and consequently they may bore users. We proposed a new reinforcement learning method specialized for the social robot, the FRAC-Q-learning, that can avoid user boredom. The proposed algorithm consists of a forgetting process in addition to randomizing and categorizing processes. This study evaluated interest and boredom hardness scores of the FRAC-Q-learning by a comparison with the traditional Q-learning. The FRAC-Q-learning showed significantly higher trend of interest score, and indicated significantly harder to bore users compared to the traditional Q-learning. Therefore, the FRAC-Q-learning can contribute to develop a social robot that will not bore users. The proposed algorithm has a potential to apply for Web-based communication and educational systems. This paper presents the entire process, detailed implementation and a detailed evaluation method of the of the FRAC-Q-learning for the first time.

**Keyword**    Communication, human-robot interaction, reinforcement learning, social robot.

## 1 Introduction

Robotics has been underpinned and applied in various fields, including social interaction [1]. Lots of robots for social interaction, or "social robot," has been released: e.g., a dog-like robot "AIBO," a seal-like robot "PARO," humanoid robots "NAO" and "Pepper" [2]. Social robots are not merely human partners, peers or pets; they can also be used as service interfaces. For example, they have been considered in the field of education as tutors, teachers, peers, and novices [3]. Social robots can also be used to cure patients [4], which is referred to as a "robot therapy."

Robotic therapy using PARO has been investigated for approximately 20 years [5]. For example, PARO has been used to treat dementia in the elderly [6]. PARO was also evaluated for home use, which showed a reduction in depressive symptoms [7]. A one-year evaluation of PARO use was also conducted [8]. Interestingly, reinforcement learning has been implemented in PARO [6]. The reinforcement learning is a type of machine learning method in which a suitable action is learned through rewards and penalties, similar to physiological learning called operant conditioning [9]. F

Reinforcement learning is often applied to social robots [10]. In robotics, reinforcement learning has been traditionally applied to control actuators in response to sensor inputs; for example, an inverse pendulum was controlled to remain standing using an autonomous railway bogie and an angle sensor. In social robots, the reinforcement learning is used to learn robot's actions via human-robot interactions. For example, an embodied robot learns actions from live human feedback using a motion-capture system [11]. Virtual social robot "Sophie" learns how to bake cake from a human partner in the "Sophie's Kitchen" [12]. A humanoid robot NAO learns via affective feedback from facial expressions to pass a red ball to a human and throw away green balls [13]. A humanoid robot Robovie II obtained suitable interaction distance between robot and human using accumulated motion and gaze meeting percentage [14]. Our previous study showed that message presentations using Q-learning and reinforcement learning attracted users more than random presentations [15].

However, reinforcement learning is not usually specialized for social robots. Unlike in the mechanical inverted pendulum problem, humans become bored with social robots over time. This implies that the suitable action for a human is not identical. To overcome this problem, we previously proposed a reinforcement learning algorithm that avoids boredom [16], and demonstrated a concrete implementation of the algorithm, named FRAC-Q-learning [17]. This algorithm has a forgetting process, in addition to the randomizing and categorizing processes. However, detailed implementation, experimental procedures, and results have not yet been presented.

The objective of this study was to clarify how FRAC-Q-learning influences participants, in contrast to traditional Q-learning. The FRAC-Q-learning was implemented in a hand-made stuffed social robot "Manami." Participants were asked to communicate with the robot. In the contrasting experiment, traditional Q-learning was applied instead of FRAC-Q-learning. After the two experiments, the participants answered a questionnaire about their interest and boredom hardness scores. The two methods are compared using a questionnaire. This study contributes to the development of social robots that will not bore users. This paper presents the entire process of FRAC-Q learning, details of its implementation, and a detailed evaluation process through an experiment for the first time.

## 2 Methods

### 2.1 Communication robot with FRAC-Q learning

Manami, as shown in Figure 1, is a prototype social robot with a new reinforcement learning algorithm called FRAC-Q-learning. Figure 2 illustrates an example of communication using Manami. Manami chooses and performs actions to interact with the users. For example, Manami plays a synthesized voice that says "Hello!" to a user. In response to this action, a user says "Hello!" to Manami. Subsequently, Manami analyzed a user's status via sensor values, for example, emotions detected by facial images from a camera. Through trial and error, Manami learned an action category that can obtain positive or very positive responses from a specific user.

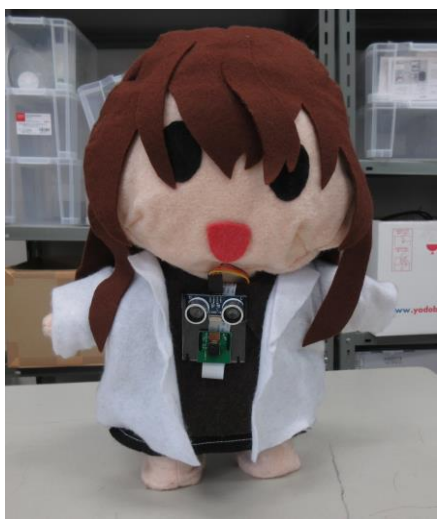

**Figure 1**  Social Robot "Manami." Her width, height, and depth are 30, 37, and 20 cm, respectively.

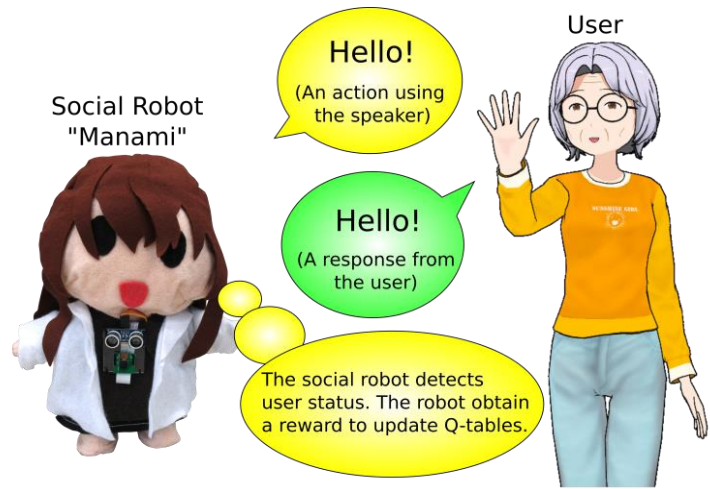

**Figure 2**   Example of communication with Manami.

### 2.1.1 Hardware of Manami

Figure 3 shows the hardware and information flow inside Manami. The bone in Manami was used to attach a single-board computer (Raspberry Pi 3 Model B+), a mobile rechargeable battery (ANKER PowerCore 10000), actuators, and sensors. Two servomotors (SG-90, TowerPro, Taiwan) attached to the shoulders of the Manami lifted the right and left hands up and down independently. In addition, a small USB speaker (MS-U1, YOUZIPPER) and a small USB microphone (MM-MCU02BK, Sanwa Supply) were installed at Manami's head. The synthesized voice files were played by a speaker using PyAudio 0.2.12. The participants' voices were recorded using the microphone. An ultrasound sensor (HC-SR04), located on Manami's chest, was used to measure the distance between Manami and the participant. A small camera (Raspberry Pi Camera v1.3) on the chest was used to categorize the participants' emotions from the facial images.

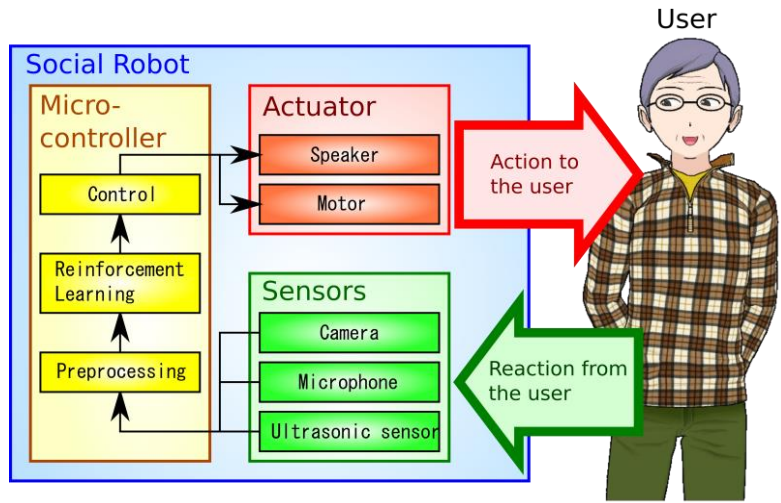

**Figure 3** Hardware and information flow inside Manami.

### 2.1.2 Software of Manami

The Raspberry Pi inside Manami runs Python 3.9.2 on a Raspberry Pi OS. It is used not only to control the actuators and sensors shown in Figure 3, but also to learn user interests based on the proposed FRAC-Q-learning. The following sections explain the forgetting, randomization, and categorization processes of the proposed FRAC-Q-learning method. FRAC-Q-learning and traditional Q-learning were implemented using the NumPy 1.23.5.

### 2.1.3 Actions and its categorization process

The actions (or behaviors) of Manami $a$ are presented to the user by controlling the servomotors and speaker. For example, Manami waves her right hand and says, "What do you want to eat?" in Japanese. These actions are categorized in the categorization process of FRAC-Q-learning; they belong to five action categories $c_a$: dancing ($c_a = 0$, three types of actions), greeting ($c_a = 1$, 5 types), questions ($c_a = 2$, 11 types), onomatopoeia ($c_a = 3$, 10 types), and jokes ($c_a = 4$, 16 types). Dancing includes waving the right, left, and both hands. The other action categories used synthesized voice files generated by VoiceBox (WhiteCUL).

### 2.1.4 State

FRAC-Q-learning has the states before an action $S_t$ and after action $S_{t+1}$. In this study, they take one of four values associated with the user's feelings estimated from the sensors (0: negative, 1: neutral, 2: positive, and 3: very positive). First, for each computation time, $S_t$ inherits $S_{t+1}$ from the previous computation time. $S_{t+1}$ is estimated after the action in the current computation time. $S_t$ is used when deciding the action category, whereas $S_{t+1}$ is important when learning a user's interests through FRAC-Q-learning.

Manami integrates multiple sensor inputs to estimate the state after action $S_{t+1}$ as follows. Once the selected action was completed, Manami began checking the sensors. Manami first detected the talk length using a microphone. The longer the talk length, the higher the score. Talk length is measured if the sound is over a threshold of sound volume within every 3 s interval. The threshold was determined by sound recorded during the initialization process. The talking scores (nSpeak) were 0, +1, and +2, corresponding to talk lengths of 0 s, 6 s, and more than 9 s, respectively. At the end of the talking, the distance between Manami and the participant was measured using an ultrasound sensor. The closer the participant was to Manami, the higher the score. Scores of the distance are -2, 0, +1, and +2 for distances (cm) $d > 100$, $100 \leq d < 40$, $40 \leq d < 20$, and $d \leq 20$, respectively. Finally, the camera on the chest captured the user's facial image using OpenCV 4.6.0.66, and the participants' facial emotions were analyzed using DeepFace 0.0.75. If the detection results are angry, sad, fear, disgust, not detected, neutral, surprise, and happy, the scores of facial emotions are -2, -2, -2, -1, 0, +1, +1 and +2, respectively. Summation of these scores in a computation time $s$ is used to estimate the current state: if $s < 0$, $0 \leq s < 1$, $1 \leq s < 3$, and $s \geq 3$, states are negative ($S_{t+1} = 0$), neutral ($S_{t+1} = 1$), positive ($S_{t+1} = 2$), and very positive ($S_{t+1} = 3$), respectively.

### 2.1.5 Rewards

A reward (penalty or punishment) is assigned after estimating $S_{t+1}$. Rewards $r_{t+1}$ of negative, neutral, positive, and very positive state are -10, -5, +5, and +10, respectively.

### 2.1.6 Q-values and Q-table

Q-values in traditional Q-learning are values which are assigned to each action in each state. The Q-value is used to estimate an action of a current state. The higher the Q-value of an action in a state, the more frequently the related action is selected in future computation time. The Q-value in traditional Q-learning is updated as follows:

$$Q(S_t, a) \leftarrow Q(S_t, a) + \alpha \left\{ r_{t+1} + \gamma \max_{a'} Q(S_{t+1}, a') - Q(S_t, a) \right\}, \quad (1)$$

where $\alpha$ is a learning rate, $\gamma$ is a discount factor, $\max_{a'} Q(S_{t+1}, a')$ returns the maximum value of all Q-value in $S_{t+1}$. Q-values can be summarized in a matrix (#state × #action), which is referred to as Q-table. The size of the Q-table increases as the number of actions increases. This implies that it requires a longer time (in steps) to learn the parameters in the Q-table. To reduce the size of the Q-table for social robot use, actions were categorized in FRAC-Q-learning.

The Q-value in FRAC-Q-learning was assigned to each action category and state. The size of Q-table is #state × #action category. Because FRAC-Q-learning has a smaller Q-table than traditional Q-learning, it can save time in learning Q-values. The Q-value in FRAC-Q-learning is computed by replacing $a$ with $c_a$ as follows:

$$Q(S_t, c_a) \leftarrow Q(S_t, c_a) + \alpha \left\{ r_{t+1} + \gamma \max_{c'_a} Q(S_{t+1}, c'_a) - Q(S_t, c_a) \right\}. \quad (2)$$

### 2.1.7 Action selection and its randomization process

Participants may become bored when the same action is repeated. The R-values were used to reduce the Q-values for which the action category was selected a few times. The R-value is computed as follows:

$$R(c_a, t_{c_a}) = \begin{cases} C_m \frac{t_s - t_{c_a}}{t_s}, & 0 \leq t_{c_a} < t_s \\ 0, & t_{c_a} \geq t_s \end{cases}, \quad (3)$$

where $t_{c_a}$ is the computation time since $c_a$ was selected, $t_s$ is the duration when the Q-value is reduced ($t_s > 0$), $C_m$ is the maximum R-value ($C_m > 0$). $t_{c_a}$ was reset ($t_{c_a} = 0$) when $c_a$ was selected, and $t_{c_a}$ increases as the computation time increased. The R-value is $C_m$ when $c_a$ is selected ($t_{c_a} = 0$). If $t_{c_a}$ was less than $t_s$, R-value decreased linearly. If $t_{c_a}$ is greater than $t_s$, the R-value is zero.

Action $a$ of state $S_t$ in Q-learning is selected based on the Q-values for each computation time. Actions with the top three Q-values were selected, with probabilities of 0.6, 0.25, and 0.13, respectively. Additionally, action $a$ was randomly selected from all the actions with a probability of 0.02.

The action category $c_a$ of state $S_t$ in FRAC-Q-learning is selected in the similar way as $a$ in traditional Q-learning. Note that $Q(S_t, c_a) - R(c_a, t_{c_a})$ are the values obtained when determining the top three values in FRAC-Q-learning. After selecting $c_a$, action $a$ in the selected action category was chosen randomly during the randomization process.

### 2.1.8 Forgetting process

In addition, the participants became bored when similar actions in the action category were repeated many times. To avoid this, Manami forgets what she learned if she had no continuous rewards from the action. In the forgetting process, the Q-table is reset (all Q-values are set to 0) when no rewards are obtained for $t_f$ times, continuously.

### 2.1.9 Flow chart

The entire FRAC-Q-learning process is illustrated in Figure 4. First, variables such as Q-tables and states were initialized. The initial Q-values were all 0. Subsequently, the number of episodes, that is, the period until the reward/penalty is given, increases. Note that the computational time represents the number of times the Q-value is calculated, which differs from the number of episodes. Manami inherits the previous state as the state before action $S_t$. Note that the initial state (1) is inherited if no previous state exists. An action is categorized in the categorization process (see section 2.1.3). In the randomization process, an action category $c_a$ is first selected before selecting an action, considering $Q(S_t, c_a) - R(c_a, t_{c_a})$. Second, action $a$ is chosen from the selected $c_a$. After presenting the chosen action $a$ to the user, the state after action $S_{t+1}$ is estimated from the sensors (see section 2.1.4). The reward or penalty $r_{t+1}$ is obtained after estimating $S_{t+1}$ (see section 2.1.5). Q-value and R values were updated (see section 2.1.6 and 2.1.7). If no reward is continuously obtained for $t_f$ times, the Q-table is reset (see section 2.1.8, the forgetting process). If a reward or penalty was given, the number of episodes increased. Note that the reward or penalty is assigned to all states of Manami, and the episode is updated every time in this study. This process is repeated to learn and adapt to user interests.

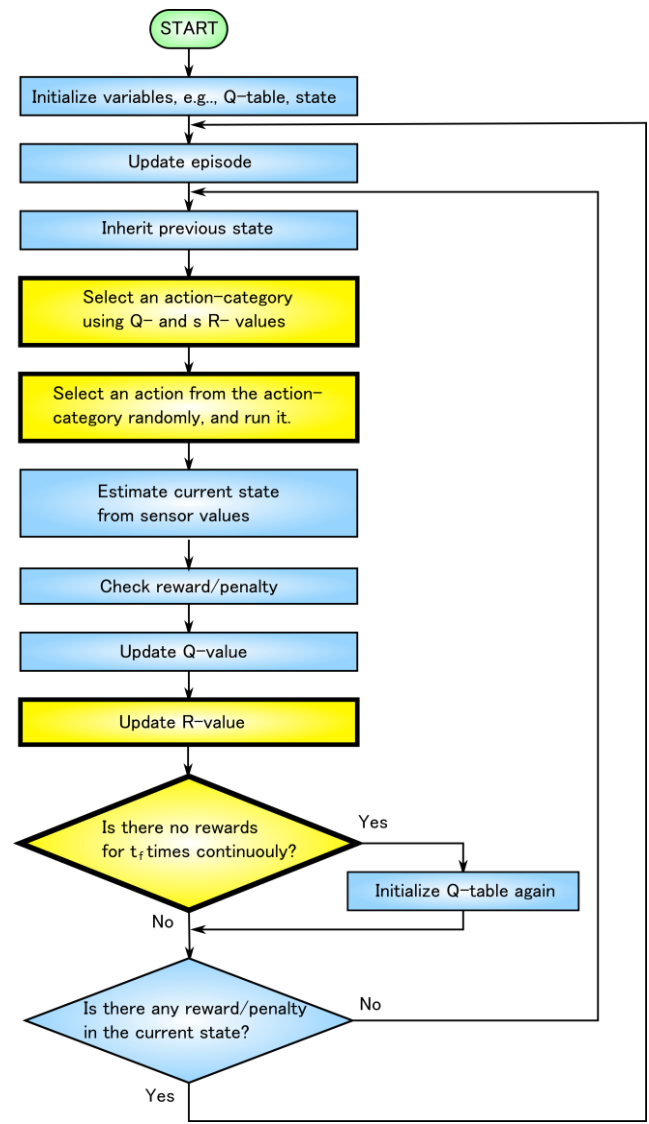

**Figure 4**  Flow chart of FRAC-Q-learning. Processes and decisions highlighted in yellow indicate unique processes of the FRAC-Q-learning. The other blue and green processes and decisions are common processes between the FRAC-Q-learning and the traditional Q-learning.

## 2.2 Experiment

This experiment was conducted to demonstrate how FRAC-Q-learning can be implemented on a social robot, and to confirm the difference in effects between FRAC-Q-learning and traditional Q-learning. FRAC-Q-learning was implemented on a stuffed social robot, Manami, with which six participants communicated. In the contrast experiment, traditional Q-learning was implemented instead of the proposed method and experiment was conducted as well. Replies written in the questionnaires were compared after the experiment.

### 2.2.1 Participants

Six healthy participants (1 female, 21.8±6.5 y.o., S01 to S06, gender in this study was based on their self-reports) took part in this experiment. All participants provided written informed consent before the experiment. This study was conducted at the National Institute of Technology, Kagawa College, Japan. This study was approved by the Ethics Committee of Shikoku Kosen Center for Innovative Technologies.

### 2.2.2 Experimental design

The experiment comprises two parts: Part A takes advantage of FRAC-Q-learning, whereas Part B uses traditional Q-learning. Manami was placed on a kitchen table and the participant was seated on a kitchen chair. Manami was located approximately 40 cm in front of the participants. The participants were asked not to wear a mask because it may influence facial emotion detection. In each part, the participants communicated with Manami for 10 min. The participants were instructed to respond to Manami after her action. The participants were allowed to rest for a few minutes between experimental sessions. The experimental order was counterbalanced to prevent order effects. After the learning, Q-tables were used to discuss the learning processes.

### 2.2.3 Questionnaires and statistical analysis

After the two experimental sessions, the participants completed a questionnaire. Participants were asked how much they were interested (very interested, interested, slightly interested, neutral, not interested slightly, not interested, not interested at all) and how much they became bored (very hard to get bored, hard to get bored, slightly hard to get bored, neutral, slightly easy to get bored, easy to get bored, very easy to get bored) by Manami in Parts A and B. To confirm the participants' motivation, they were asked to write about Manami's actions as much as they memorized them and were asked to guess the learning mechanism. The replies to the questionnaire were statistically analyzed using Welch's test.

### 2.2.4 Parameters

The parameters used in this study are listed in Table 1. $\alpha$ and $\gamma$ are common in both the experimental parts. $t_f, C_m,$ and $t_s$ are unique parameters of FRAC-Q-learning.

Table 1　Parameters of FRAC-Q-learning and Traditional Q-learning in the experiment.

| Variable | FRAC-Q-learning | Traditional Q-learning |
|---|---|---|
| $\alpha$ | 0.9 | 0.9 |
| $\gamma$ | 0.5 | 0.5 |
| $t_f$ | 10 | - |
| $C_m$ | 15 | - |
| $t_s$ | 3 | - |

# 3 Results

Figure 5 shows interest scores. The higher the interest score, the more interested the participant was in the system. Interest score of experiment using FRAC-Q-learning was 2.17 ±0.41, whereas that using traditional Q-learning was 0.67 ± 1.75. Welch's test showed a significant trend in interest scores ($p = 0.09$). FRAC-Q-learning tended to show a higher degree of interest than traditional Q-learning.

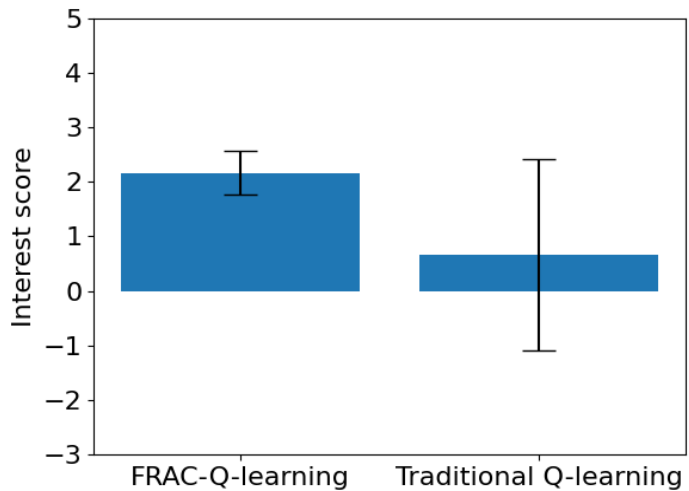

Figure 5　Interest score.

Figure 6 shows the boredom hardness scores. The higher the boredom hardness score gets, the harder for the system to bore users. The boredom hardness score of the FRAC-Q-learning was $1.17 \pm 1.72$, whereas that of the traditional Q-learning was $-1.8 \pm 0.98$. The FRAC-Q-learning showed a significantly higher boredom hardness score than the traditional Q-learning ($p < 0.01$). These results imply that it is more difficult for FRAC-Q-learning to bore the participants.

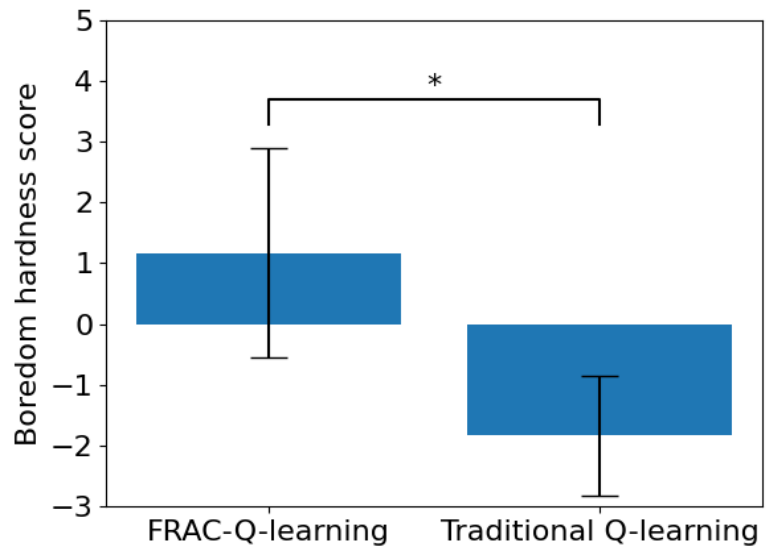

Figure 6  Boredom hardness score.

Figure 7 shows the heatmaps of the Q-tables. The vertical axis shows the state $S_t$ (0: negative, 1: neutral, 2: positive, and 3: very positive), whereas the horizontal axis shows the action category or action. Traditional Q-learning has a larger untrained green area. This implied that additional training episodes were required. States 0 to 2 had high Q-values in some actions or action categories, whereas state 3 had non-zero values only for subjects 1 and 3. This means that it is difficult to obtain State 3 using the current sensor inputs or scoring methods.

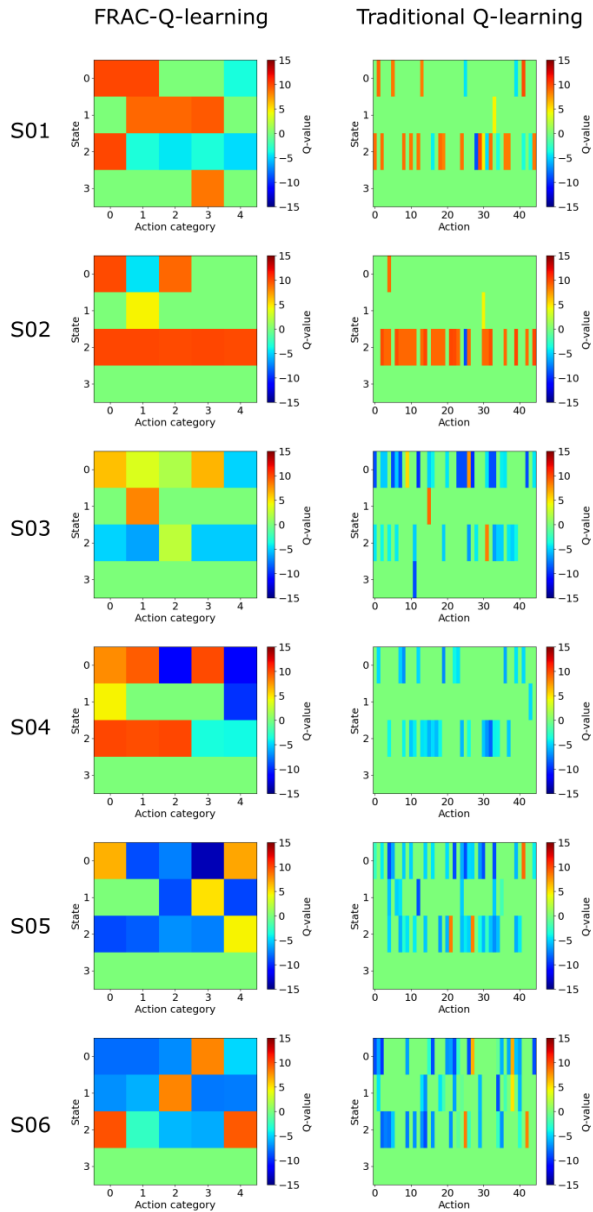

**Figure 7** Heatmaps of Q-tables.

Figure 8 shows the number of talking (nSpeak) in each session. The vertical axis shows the number of times the participants spoke for, and the horizontal axis represents the learning steps. Four of the six showed that FRAC-Q-learning had more talking than traditional Q-learning. This means that the participants spoke back to the system in response to the robot's actions more frequently in FRAC-Q-learning.

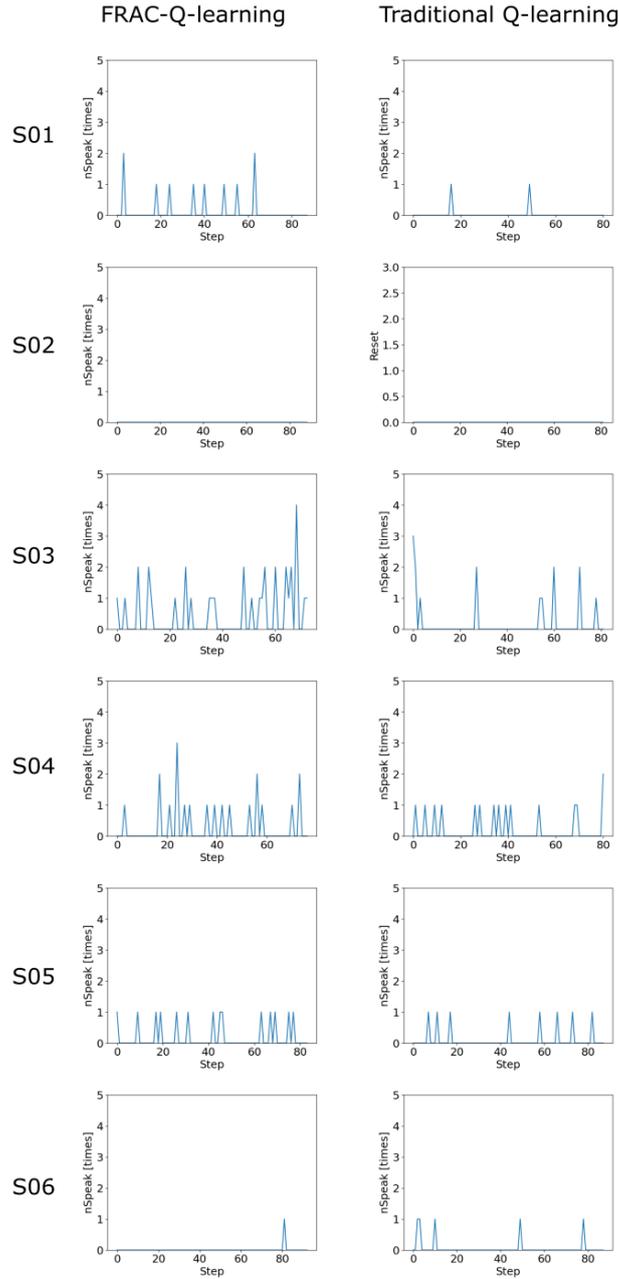

**Fig. 8**  The talking scores (nSpeak) during experiments.

# 4 Discussion

Toward developing a social robot that does not easily bore users, this study evaluated the proposed FRAC-Q-learning, an extension of traditional Q-learning that has a forgetting process in addition to randomizing and categorizing processes. To reveal the effect of FRAC-Q-learning, it was implemented in a social robot, Manami, and compared with traditional Q-learning. FRAC-Q-learning tended to show higher interest scores and significantly higher boredom hardness scores than traditional Q-learning. These results imply that FRAC-Q learning can be used to develop a social robot that will not bore users.

The boredom hardness score of FRAC-Q-learning was significantly higher than that of traditional Q-learning. A previous study also reported boredom using the reinforcement learning [18]. Although boredom was not explicitly measured or evaluated, it is an important issue in the development of a social robot. Boredom was evaluated using a questionnaire. However, it would be stricter if boredom could be measured using biological signals such as electroencephalography or electrocardiography. Currently, no reliable biological signals have been applied in social robot studies.

Significant trends in interest scores were observed between the two methods. In our previous study, a message presentation system using traditional Q-learning was significantly better than that using random presentation [15]. The differences in the current study was not significant, because both algorithms are a kind of Q-learning.

Future applications of FRAC-Q learning include web communication systems, human learning systems, web advertising systems, and robot therapy. This study demonstrated the implementation of an algorithm for social robots. However, the robot does not need to have its own body, except for special uses such as touch interaction. A 3D avatar can be used as an interface for reinforcement learning instead of a physical robot. Although there are some differences between real and virtual bodies, a virtual body is convenient because it can be presented via the display of computers or smartphones. FRAC-Q-learning can also be applied to learning systems. For example, humans can learn human-to-human communication skills through human-to-robot communications. FRAC-Q-learning have a potential to present difficult questions by applying rewards when the answer is not good, and penalties (punishments) when the answer is good. The proposed algorithm may learn an efficient web advertisement category: rewards are given when mouse clicks or mouseover events are detected, whereas penalties are applied when no response is detected until other web pages are opened. Finally, the proposed algorithm may be applied to robot therapy.

Reinforcement learning contributes to robot therapy; however, clear scientific evidence has not yet been obtained. Previously, a social robot "PARO" showed the effect of robot therapy. However, no element-wise experiments have been conducted. PARO contains many factors; therefore, this study could not identify the most effective factor. To clarify the effect of reinforcement learning on robotic therapy, a special experimental design is required.

Many issues still need to be resolved toward developing better social robots. For example, the maximum number of actions is limited by the current implementation. However, the number of actions can be increased to infinity using generative artificial intelligence and voice synthesizers [19]. In other words, the actions in an action category can be generated automatically. Communication in a noisy environment, for example in a car, party, or poster presentation session, is also difficult. Traditionally, these problems have been addressed through blind-source separation using independent component analysis [20]. This study focused on communication between a robot and a human. This situation can be extended to communication between a robot and many humans, between robots, or between robots and humans. It may be possible for robots and animals to communicate [21]. This study also limited the situation in which the robot initiates communication with a human; however, robots need to respond to human actions.

## 5. Conclusion

To clarify how FRAC-Q-learning influences participants, it was implemented in a hand-made stuffed social robot, Manami, and compared with traditional Q-learning through a human-robot interaction experiment. FRAC-Q-learning showed significantly higher boredom hardness scores than traditional Q-learning. These results imply that FRAC-Q-learning can contribute to the development of social robots that can avoid user boredom. This algorithm has a potential to apply to Web communication systems, learning systems for humans, and Web advertising systems.

## Acknowledgements

This work was supported in part by the Tateisi Science and Technology Foundation (Grant Number 2221008).

## Disclosure of conflicts of interest

The author declares no conflicts of interest associated with this manuscript.